\documentclass[sigconf]{aamas}

\usepackage{balance} 
\usepackage{amsmath,amsfonts}
\usepackage{eqparbox}
\usepackage{algorithm}
\usepackage{subcaption}
\usepackage{booktabs} 
\usepackage{siunitx}  
\usepackage{algpseudocode}
\usepackage{multirow}
\usepackage{booktabs}
\usepackage{color, soul}
\usepackage{xcolor}
\usepackage{adjustbox}
\usepackage[ font=footnotesize]{caption}
\usepackage{hyperref}
\newtheorem{remark}{Remark}

\setcopyright{none}

\acmDOI{}
\acmPrice{}
\acmISBN{}
\acmConference[]{\textbf{$^*$Indicates equal contribution}}{This research is supported by funding from the Dutch Research Council NWO-NWA, within the “Acting under Uncertainty” (ACT) project (Grant No.NWA.1292.19.298)}{$^{\dagger}$Work done partially while at Delft University of Technology. Author acknowledges partial support from UKRI grant EP/W002949/1}
\copyrightyear{}
\acmYear{}

\title[Predictable and Robust Multi-Agent Coordination]{Predictability Awareness For Efficient and Robust\\ Multi-Agent Coordination}

\author{Roman Chiva Gil}
\affiliation{
  \institution{Delft University Of Technology}
  \city{Delft}
  \country{Netherlands}}
\email{R.ChivaGil@student.tudelft.nl}

\author{Daniel Jarne Ornia*$^{\dagger}$}
\affiliation{
  \institution{University of Oxford}
  \city{Oxford}
  \country{United Kingdom}}
\email{daniel.jarneornia@cs.ox.ac.uk}

\author{Khaled A. Mustafa*}
\affiliation{
  \institution{Delft University Of Technology}
  \city{Delft}
  \country{Netherlands}}
\email{k.a.mustafa@tudelft.nl}

\author{Javier Alonso Mora}
\affiliation{
  \institution{Delft University Of Technology}
  \city{Delft}
  \country{Netherlands}}
\email{j.alonsomora@tudelft.nl}

\begin{abstract}
To safely and efficiently solve motion planning problems in multi-agent settings, most approaches attempt to solve a joint optimization that explicitly accounts for the responses triggered in other agents. This often results in solutions with an exponential computational complexity, making these methods intractable for complex scenarios with many agents. While sequential predict-and-plan approaches are more scalable, they tend to perform poorly in highly interactive environments. This paper proposes a method to improve the interactive capabilities of sequential predict-and-plan methods in multi-agent navigation problems by introducing predictability as an optimization objective. We interpret predictability through the use of general prediction models, by allowing agents to predict themselves and estimate how they align with these external predictions. We formally introduce this behavior through the free-energy of the system, which reduces (under appropriate bounds) to the Kullback-Leibler divergence between plan and prediction, and use this as a penalty for unpredictable trajectories. The proposed interpretation of predictability allows agents to more robustly leverage prediction models, and fosters a ‘soft social convention’ that accelerates agreement on coordination strategies without the need of explicit high level control or communication. We show how this predictability-aware planning leads to lower-cost trajectories and reduces planning effort in a set of multi-robot problems, including autonomous driving experiments with human driver data, where we show that the benefits of considering predictability apply even when only the ego-agent uses this strategy. The code and experiment videos can be found in the following page: \href{https://romanchiva.github.io/PAProjectPage/}{https://romanchiva.github.io/PAProjectPage/}
\end{abstract}


\newcommand{\BibTeX}{\rm B\kern-.05em{\sc i\kern-.025em b}\kern-.08em\TeX}

\begin{document}


\pagestyle{fancy}
\fancyhead{}


\maketitle 


\section{Introduction}

Many modern robotics applications involve autonomous agents navigating multi-agent environments where they will be required to interact with humans and other robots without full
knowledge or extensive communication capabilities \cite{wang_social_2022}. This involves planning trajectories in a complex system governed by a mix of rational and non-rational, stochastic and possibly game
theoretic behaviors. To achieve safe and efficient interactions, agents need to reason about each other and coordinate. However, this poses critical challenges due to the high uncertainty associated with estimating other agents’s objectives \cite{fisac_hierarchical_2018} and a computational complexity that renders problems intractable for more than a handful of agents. 

\renewcommand{\thefootnote}{}
\footnotetext{\textbf{$^*$Equal contribution}.\\ This research is supported by funding from the Dutch Research Council NWO-NWA, within the “Acting under Uncertainty” (ACT) project (Grant No.NWA.1292.19.298). $^{\dagger}$Work done partially while at Delft University of Technology. Author acknowledges partial support from UKRI grant EP/W002949/1}
\renewcommand{\thefootnote}{\arabic{footnote}}

Receding Horizon Trajectory Optimization allows for flexible and anticipative planning while ensuring compliance with \emph{e.g.} safety constraints in multi-agent navigation problems. However, planning a trajectory that explicitly accounts for interactions among agents generally requires solving a joint optimization problem. A variety of joint planning methods can be found in literature, \emph{e.g.} \cite{Huang_dtpp_2024, Chen_2023}, of which game theoretic approaches best capturing agent interaction complexities \cite{wang_social_2022}. By modeling other agents as rational actors, game theoretic approaches cast the joint optimization as a constrained dynamic game and seek to find equilibrium solutions. Although this often results in stable and coordinated interactions \cite{fisac_hierarchical_2018, cleach_algames_2020}, game theoretic approaches suffer from the curse of dimensionality, as the planning complexity grows exponentially with the number of agents \cite{schwarting_planning_2018}. Additionally, modeling other agents as rational is a strong assumption which will not hold in practice, especially when interacting with human agents \cite{colman_cooperation_2003, bossaerts_computational_2017}. 

Alternatively, predict-and-plan approaches scale well with number of agents, however they tend to perform poorly in interactive environments. By separating prediction and planning, the problem simplifies to a single-agent collision avoidance problem with dynamic obstacles \cite{brito_model_2020, de_groot_scenario-based_2023}. The accuracy of the prediction model limits how well agents can coordinate. A system of interacting agents is highly complex, making it difficult to predict the diversity of possible futures, especially when considering interactions. This can lead to ambiguous predictions, making agents unable to anticipate their environment, and thus have to re-plan more often or engage in riskier behaviors \cite{wang_social_2022}.

Ideally, every agent in the environment would be able to accurately anticipate surrounding agents' future trajectories allowing for efficient and safe interaction. Sequential planning agents use prediction models to avoid collisions with others, however, this fails to acknowledge that surrounding agents also hold predictions about the ego-agent, and plan their trajectory based on these predictions. Unless the optimal avoidance strategy falls within the range of predicted behaviors, other agents will react to the unexpected avoidance strategy by modifying their own trajectory. 
To mitigate this issue, we propose the following: in the same way a prediction model is used to predict other agents, the ego-agent can use it to approximate how other agents expect it to behave. This information can be used in planning to introduce a penalty for trajectories other agents will find surprising, bringing the optimal trajectory closer to the expectation surrounding agents hold. 
Accounting for predictability in this way mirrors the principle of free-energy minimization in active inference \cite{sajid2021active} (and control systems \cite{theodorou2010generalized}), where an agent not only seeks to maximize reward but also aims to minimize the discrepancy between some prediction model and observations. In multi-agent interactions \cite{hyland2024freeenergy}, agents hold probabilistic beliefs about the behavior of others, and the accuracy of these beliefs is directly influenced by the agent's own actions. By minimizing free energy, the agent balances actions that reduce uncertainty and confirm its internal model of the world with those that maximize reward. This approach ensures that the agent's behavior is not only goal-directed but also aligned with maintaining coherent and accurate beliefs about the surrounding agents.


\subsection{Contribution}
We explore how sequential planning agents can improve their coordination capabilities by accounting for the predictability of their planned trajectories. When a group agents accounts for predictability, they are able to foster a ‘soft social convention’ dictated by the prediction model which results in a decrease of uncertainty about the environment for all agents in the group. This helps agents resolve coordination problems without having to explicitly model interactions.  Formally, the contribution of this paper is threefold:

\begin{enumerate}
    \item  We exploit ideas on free-energy to formulate a cost function that uses feedback from a prediction model to include predictability as an objective and analyze how this cost function can be integrated with a planner.

    \item We provide results showing how our predictability awareness mechanism leads to `soft social conventions' forming-based interaction strategies encoded in prediction models for multi-robot navigation problems. This allows agents to achieve smoother coordination by improving the effectiveness of prediction models in interactive environments.
    
    \item Accounting for predictability causes agents to adopt \emph{social norms} and pro-social behaviors encoded in learned prediction models, allowing to more closely mimic experts' behaviors without needing cost function learning. We provide evidence for these behaviors in an experiment where an agent interacts with human drivers in scenarios from the Waymo Open Motion Dataset.

\end{enumerate}
\section{Related Work}
\label{Related Work}
\paragraph{\textbf{Integration of Prediction Models and Planners}} Trajectory prediction has significantly advanced in recent years, particularly with the development of transformer-based generative models capable of producing interaction-aware joint trajectory predictions, \emph{e.g.} \cite{eccv2020, scept2022, huang_survey_2022}. While these models show impressive performance in open-loop evaluations, integrating them with planners in highly interactive settings remains challenging \cite{hagedorn_rethinking_2023}. Effective interactive planning often necessitates joint prediction and planning. Additionally, the planner often requires some form of learned cost function \cite{huang_differentiable_2023}. Otherwise, if the behavior of the expert significantly differs from the expert in the training data, this will throw the model out of distribution yielding low quality predictions.

Many studies have focused on developing ego-conditioned prediction models \cite{ngiam_scene_2022}; however, their integration with planners faces obstacles primarily due to computational complexity. For instance, in \cite{Huang_dtpp_2024} Tree Policy Planning (TPP) has been employed to generate an initial set of partial trajectories, which condition the prediction model and create a scenario tree. This tree is evaluated using a cost function combining designed and learned features to identify and expand promising scenarios, efficiently allocating computational resources. A novel approach by \cite{Chen_2023} leverages unconditioned prediction models to provide initial estimates of other agents' trajectories, capitalizing on the models' ability to predict general intentions accurately while acknowledging their limitations in capturing short-term interaction details. This approach optimizes the ego and agent trajectories together, minimizing disturbances from the initial agent paths and utilizing homotopy classes to ensure diversity and avoid local minima. Instead of conditional prediction models, some methods develop fully differentiable stacks \cite{karkus_diffstack_2022,huang_differentiable_2023} enabling gradient backpropagation through the planner, which allows for combined prediction model fine-tuning and cost function learning aligned with expert behavior in the training data. While avoiding the joint optimization, our approach links prediction and planning without the need for retraining or fine-tuning by including a term in the cost function that helps guide the agent's behavior to not compromise its predictions. This allows for maintaining flexibility in selecting prediction models and planner combinations while being compute-efficient.

\paragraph{\textbf{Predictability and Legibility of Motion}} In the field of Human-Robot Interaction, legibility and predictability of motion have been studied to improve coordination by designing agent behaviors that clearly communicate intention and avoid surprising observers \cite{dragan_legibility_2013}. Often both objectives overlap \cite{bhatt_efficient_2023}. Traditional formulations of this problem are not well suited for receding horizon applications as they optimize over complete trajectories and rely on utility-based analytical models of observer expectations \cite{dragan_integrating_2014}. Additionally, the observer is modeled as inactive, thus having no influence on the planning agent. This assumption breaks down in multi-agent navigation where the observer and the agent share the workspace and influence each other. Several works have explored the adaptation of these concepts to an interactive multi-agent context. In single agent RL settings, the impact of predictability objectives has been recently studied in \cite{ornia2023predictable}. In multi-agent scenarios, \cite{bastarache_legible_2023} define dynamic goal regions around neighboring agents and optimize for reduced uncertainty about the collision avoidance strategy. \cite{geldenbott_legible_2024} show how increasing action penalties at later horizon steps causes agents to more rapidly demonstrate their avoidance strategy. This accelerates intent inference giving agents better anticipation. \cite{brudigam_legible_2018} defines hand-crafted legibility costs for planning in highway driving. These methods are often designed to target a specific type of observer model. In contrast, our approach minimizes a predictability surrogate that allows modeling the observers with an arbitrary prediction model choice. 

\section{Trajectory Planning}
The general optimization problem for a single-agent in stochastic motion planning can be formulated as follows:
\begin{subequations}
\begin{alignat}{2}
\min_{\boldsymbol{u} \in \mathbb{U}, \boldsymbol{x} \in \mathbb{X}} \quad & \sum_{k=0}^{K-1}J_k(\boldsymbol{x}_k, \boldsymbol{u}_k) + J_K(\boldsymbol{x}_K)\\
\textrm{s.t.} \quad & \boldsymbol{x}_0 = \boldsymbol{x}_{\text{init}},\\
 & \boldsymbol{x}_{k+1} = f(\boldsymbol{x}_k, \boldsymbol{u}_k), \quad k = 0,...,K-1 \label{eq1b}  \\
 & \mathbb{P} \left[C(\boldsymbol{x}_k, \boldsymbol{\delta}_k^o), \forall o \right] \geq 1-\epsilon_k, \forall k, \label{CC} 
\end{alignat}
\label{ProblemFormulation}
\end{subequations}

\noindent where $\boldsymbol{u} = \{\boldsymbol{u}_0,...,\boldsymbol{u}_{K}\} \in \mathbb{U}$ are the system inputs subject to input constraints, $\boldsymbol{x}_k \in \mathbb{X} $ denotes the states of the robot, $f(\cdot)$ corresponds to the nonlinear system's dynamics, $J_k(\boldsymbol{x}_k, \boldsymbol{u}_k) \geq 0$  is the cost function specifying performance metrics, and $K$ is the length of the planning horizon. In this formulation, $C(\cdot)$ is the collision avoidance constraint, and $\boldsymbol{\delta}_k^o$ is the uncertain position of obstacle $o$ at stage $k$ obtained through a prediction model $\mathcal{P}(\mathbf{X})$ that takes into account the concatenated states of all agents in the scene. The chance constraint in Eq. \eqref{CC}, guarantees that the probability that the robot collides with the dynamic obstacle is below a specified threshold $\epsilon_k$.

In a game theoretic setting where all agents are controlled by a centralized planner, the problem reduces to solving a joint optimization program over all agents and all possible trajectories, such that from the set of joint trajectories that satisfy the constraints, the agents execute the optimal ones. This naturally carries high computational complexity, access to some centralized controller, and full information assumptions. Consider instead the case where $N$ (interactive) agents solve the optimization problem \eqref{ProblemFormulation} independently and use model $\mathcal{P}\left(\mathbf{X}\right)$ to predict each other (and thus estimate the probabilities of constraint satisfaction). Agents can then query the prediction model to observe the predictions others have about them. Our method reduces to the following intuition: Agents can use this information to shift their behaviors towards the distribution coming out of the prediction model. This `closes the loop' on prediction errors, intuitively improving the planning problem in two ways. First, inducing implicit decentralized coordination: an ideal situation is one where all agents act following the model-predicted distribution, and this distribution perfectly optimizes the cost of each agent. Second, it `robustifies' the prediction model \emph{a posteriori}: once the model has been trained on offline data, agents actively shift their plans towards the predicted distributions, collectively reducing prediction errors and widening the space of suitable prediction models for a given problem.

\section{Proposed Method: Free Energy as a Predictability Surrogate}
\label{Methodology}

\subsection{Derivation of a predictability aware cost function}

Our objective is to design a framework that allows agents to trade off predictability with progress toward the goal. If we define an agent's optimal trajectory distribution as $\mathcal{Q}^*$, in the best-case scenario, an agent's optimal trajectory distribution aligns with the predictions held by other agents. This alignment allows the agent to minimize its own cost while avoiding any disruption or interference with the trajectories of surrounding agents. In this case, no trade-off needs to be performed, however, deviations from this ideal scenario are to be expected. To formalize this as a planning objective, agents should seek to minimize the cost of trajectories sampled from their corresponding prediction in $\mathcal{P}\left(\mathbf{X}\right)$. Drawing inspiration from the path integral control  derivation in \cite{williams_information_2017}, we begin by defining the free energy of a trajectory distribution: 
\begin{equation*}
\mathcal{F}(S, \mathcal{P}, \lambda) = -\lambda \log(\mathbb{E}_{\boldsymbol{x} \sim \mathcal{P}}[\text{exp}(-\frac{1}{\lambda}S(\boldsymbol{x}))]),
\end{equation*}
where $S$ is a state cost function that \emph{represents some (trajectory planning) objective}, $\mathcal{P}$ denotes a prediction distribution, $x$ is a trajectory sampled from $\mathcal{P}$, and $\lambda$ represents the inverse temperature controlling the strictness of the efficiency criterion. This control theoretic free energy can be interpreted as a measure of how efficient a prediction distribution is at minimizing cost $S$. The free energy is minimized by pushing $\mathcal{P}$ as close as possible to $\mathcal{Q}^*$. 

The free energy as defined so far is a function of prediction distribution $\mathcal{P}$, however, agents won't plan trajectories by sampling from $\mathcal{P}$. Instead, we define $Q$ as a trajectory distribution an agent has control over. Let the states $\boldsymbol{x} = \{\boldsymbol{x}_0, ..., \boldsymbol{x}_K\}$, which the ego-agent occupies along its planned trajectory $\tau_{0, K}$, be represented as narrow Gaussians $q(\boldsymbol{x}_k)$  with mean $\boldsymbol{x}_k$ covariance $\Sigma_k$:
\begin{equation}
\begin{aligned}
\tau_{0:K} &= \{q(\boldsymbol{x}_k)\}_{k=0}^{K}, \\
q(\boldsymbol{x}_k) &= \mathcal{N}(\boldsymbol{x}_k, \Sigma_k).
\end{aligned}
\label{DistribTraj}
\end{equation}

By applying an expectation switch, these distributions can be incorporated into the free energy definition, making it a function of the agent's plan,

\begin{equation}
    \mathcal{F}(S,\mathcal{P}, \lambda) =  -\lambda\log(\mathbb{E}_{\boldsymbol{x} \sim \mathcal{Q}}[\text{exp}(-\frac{1}{\lambda}S(\boldsymbol{x}))\frac{p(\boldsymbol{x})}{q(\boldsymbol{x})}]),
\label{ExpectationSwitch}
\end{equation}
where $p$ is the density function of the prediction. By concavity of the logarithm and Jensen’s inequality, 

\begin{equation*}
    \mathcal{F}(S, \mathcal{P}, \lambda) \leq -\lambda\mathbb{E}_{\boldsymbol{x} \sim \mathcal{Q}}[ \log(\text{exp}(-\frac{1}{\lambda}S(\boldsymbol{x}))) +\log(\frac{p(\boldsymbol{x})}{q(\boldsymbol{x})})].
\end{equation*}

Finally, using the definition of Kullback-Leibler Divergence and simplifying,
\begin{equation}\label{eq:free_energy_bound}
    \mathcal{F}(S,\mathcal{P}, \lambda) \leq \mathbb{E}_{\boldsymbol{x} \sim \mathcal{Q}}[S(\boldsymbol{x})] +\lambda \mathbb{KL}(q(\boldsymbol{x})||p(\boldsymbol{x})),
\end{equation}
where $\mathbb{KL}$ denotes the KL-Divergence. The right-hand side provides an upper bound on the free energy, and one can minimize this instead of the free energy. It resembles a standard control objective, and the terms allow for good conceptual understanding of the effect they have: A \textbf{Performance Cost} and \textbf{Predictability Cost} respectively, which penalizes agents for acting unpredictably. Using this newly found expression as a stage cost, we can craft the following cost function as a stage cost for a planning problem:

\begin{equation*}
    J(\tau_{0:K}) = \sum_{k=0}^{K} J_{k}(\boldsymbol{x}_k,\boldsymbol{u}_k) + \lambda \mathbb{KL}(q(\boldsymbol{x}_k)||p(\boldsymbol{x}_k)),
\end{equation*}


where we implicitly assume $J_k$ to be composed by some state cost $S$ and some control action cost. Minimizing this cost function allows agents to trade off predictability and progress toward the goal by means of the free energy, and $\lambda$ can be selected to control how much weight is assigned to predictability during planning.


\begin{remark}
We can emphasize now the intuition behind using the free energy as a way of incorporating predictability into optimal control. Eq. \eqref{eq:free_energy_bound} is minimized precisely when $\mathcal{Q}=\mathcal{Q}^*=\mathcal{P}$. That is, the trajectory distribution executed is exactly the optimal cost trajectory distribution, and this matches the predicted distribution. Under this condition, the agent is behaving without surprising external observers and simultaneously obtaining optimal cost in its objective.
\end{remark}

\subsection{Integration with a Planner and Practicalities}

The KL-Divergence expression only has closed form solutions for a restricted set of distributions, thus to accommodate arbitrary distributions, the KL divergence term will often need to be evaluated through sampling with $\mathcal{Q}$ the candidate trajectory distribution and $P$ the prediction distribution from $\mathbb{KL}(\mathcal{P} \| Q) = \mathbb{E}_{\boldsymbol{x} \sim P} \left[\log \frac{\mathcal{P}(\boldsymbol{x})}{Q(\boldsymbol{x})}\right]$. Since sampling is required to evaluate the cost function, this could render the use of gradient based MPC unfeasible for real time planning, additionally prediction distributions $Q(\boldsymbol{x})$ may not always be differentiable. We find it is more practical to rely on sampling based MPC approaches, as they don't require a differentiable cost function and computations can be easily parallelized to handle large numbers of samples even when it is computationally expensive to evaluate the cost function. In our experiments, Section \ref{Experiments}, we rely on an Model Predictive Path Integral (MPPI) control method \cite{williams_information_2017}. 

Another consideration is that predictions about an agent's future are updated as new observations are received. For this reason, it is most effective to focus on early horizon time-steps when evaluating a plan's predictability. Thus we propose to discount the predictability cost along the horizon with factor $\gamma$ to account for uncertainty about future predictions: 
\begin{equation}
    J(\tau_{0:K}) = \sum_{k=0}^{K} J_{k}(\boldsymbol{x}_k,\boldsymbol{u}_k) + \gamma^k \lambda \mathbb{KL}(q(\boldsymbol{x}_k)||p(\boldsymbol{x}_k)).
    \label{eq:cost}
\end{equation}

\begin{remark}
It should be noted that our method is agnostic to the choice of the planner. However, in case MPPI is used as the planner, similar to \cite{biased_mppi}, our approach can be interpreted as leveraging the distribution of the prediction model as an ancillary controller to influence the MPPI sampling process.
\end{remark}
\section{Experiments}
\label{Experiments}
We present here the experiments carried out to validate our method. The first experiment investigates how accounting for predictability affects an individual agent's behavior, comparing the results with other observer-aware planning approaches. The second experiment examines the impact of predictability within a group of agents, focusing on swapping tasks in an open environment to give insight without external environmental influences. In the third experiment, we explore a practical driving scenario, demonstrating how predictability-aware agents can better coordinate and utilize prediction models. We also observe that agents indirectly exhibit expert-like behaviors, such as following social norms, without explicitly encoding them in the planner. Finally, the fourth experiment explores this direction further by testing interactions with recorded human driver data using a state-of-the-art prediction model, showing that predictability-aware agents achieve safer trajectories as a result of more closely mimicking human behavior.

\subsection{Planner}
For all experiments in this section, we use a sampling-based planner, namely Model Predictive Path Integral (MPPI) control \cite{pezzato2023samplingbased}, based on the methodology presented in \cite{williams_information_2017}. MPPI places no restrictions on dynamics model or cost function and converges well toward optima with a moderate amount of samples \cite{williams_model_2017}. Given a nominal control sequence as an initial guess, MPPI applies Gaussian noise at each step to generate a set of $M$ control sequence samples. It then uses a state transition function $f(\cdot)$ to simulate their corresponding $M$ state trajectories. Each of the resulting state trajectories is evaluated based on the cost defined in \eqref{eq:cost}, resulting in a total sample cost $J_m$. Once $J_m$, $\forall m \in [1,...,M]$ is computed, importance sampling weights, $w_m$, can be calculated as:
\begin{equation*}
w_m = \frac{1}{\eta} \exp\left(-\frac{1}{\lambda} (J_m - J_{\text{min}})\right), \quad \sum_{m=1}^{M} w_m = 1,
\end{equation*}
where $J_{\text{min}}$ is the minimum sampled cost, $\eta$ is a normalization factor and $\lambda$ is a controlling parameter that controls the width of the weight distribution. These weights prioritize lower-cost trajectories. The optimal control sequence \( U^* \) is then calculated as the weighted sum of all sampled control sequences:
\begin{equation*}
U^* = \sum_{m=1}^{M} w_m U_m,
\end{equation*}
where we use $U_m=\{\boldsymbol{u}_0,\boldsymbol{u}_1,...,\boldsymbol{u}_K\}$ to denote the $m$-th sampled control sequence. It is common to use a time-shifted version of $U^*$ to warm-start the sampling strategy at the next time-step. 

\subsection{Metrics}
As a proxy to measure coordination, we propose the use of planning effort. Planning effort is a metric taken from \cite{carton_measuring_2016} to quantify how much trajectories deviate from an initial estimate. The authors point out this serves as a proxy for how well the agent is able to anticipate the evolution of its surroundings. We adapt planning effort for receding horizon tasks with the following formulation:

\begin{equation}
\begin{aligned}
    &PE( \xi_{0:T}) = \frac{1}{T-1}\sum_{t=0}^{T-1} MSE(\tau_t, \tau_{t+1}), \ \text{with}\\
    &MSE(\tau_t, \tau_{t+1}) = \sum_{k=0}^K \| \boldsymbol{x}_k^t - \boldsymbol{x}_k^{t+1}\|,
    \end{aligned}
\end{equation}

\noindent where $T$ and $K$ denote the total time duration of the simulation and planning horizon length respectively. $\xi_{0:T}$ denotes the set of all the plans along a trajectory $\xi_{0:T}=\{ \tau_0, \tau_1, \ldots, \tau_{T-1}\}$: with $\tau_t$ the plan at time-step $t$. $\boldsymbol{x}_k^t$ represents the state at horizon step $k$ for the plan $\tau_t$. In this context, planning effort measures, on average, the magnitude of an agent's plan update per time-step. Generally, for a given task, a more accurate prediction model corresponds to a lower planning effort.

\subsection{Single Agent Experiments}
\begin{figure}[h!]
\begin{subfigure}{0.225\textwidth}
  \includegraphics[width=\textwidth]{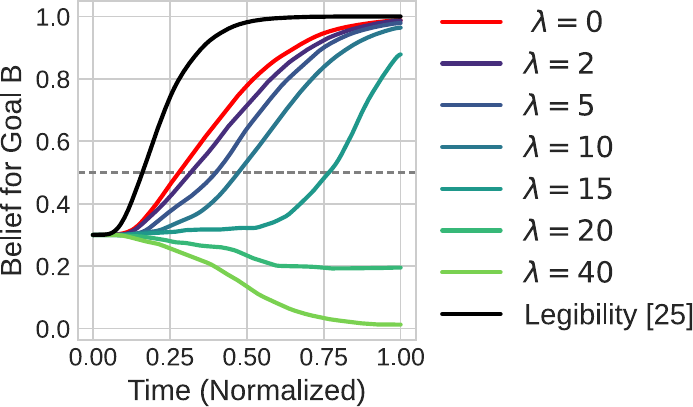}
  \captionsetup{justification=centering}
  \caption{Belief updates over trajectories.}
  \label{fig:BeliefUpdates}
\end{subfigure}%
\begin{subfigure}{0.225\textwidth}
  \includegraphics[width=\linewidth]{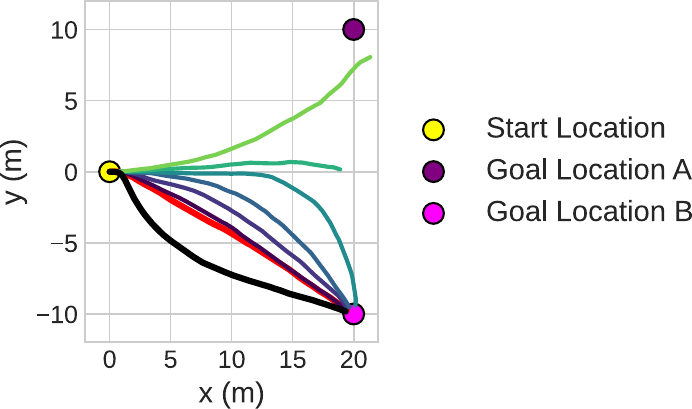}
  \captionsetup{justification=centering}
  \caption{Trajectories for different $\lambda$.}
  \label{Trajs}
  \end{subfigure}%
\caption{Figure \ref{fig:BeliefUpdates} shows that increasing $\lambda$ effectively decrease the belief update rate for the observer. In Figure \ref{Trajs}, the nominal trajectory is rendered in red. Given the observer holds mistaken initial beliefs about the robot's goal, we observe that increasing the predictability score $\lambda$ results in trajectories that are more compliant with the observer's expectation.}
\label{traj}
  \end{figure}

\paragraph{\textbf{Experiment Objective}}
In this experiment, we present a single agent interacting with a hand-crafted multi-modal prediction model, serving as a model of an observer's expectation. This is a benchmark task used by previous works on legibility and predictability \cite{dragan_integrating_2014}, \cite{mcpherson_efficient_2021} to provide clear insight into the relationship between predictability and the agent's intrinsic motivation.
\paragraph{\textbf{Setup}}
Consider an environment with two possible goals: $\mathcal{G} = \{A:[20,10], B:[20,-10] \}$. The robot starts at position $\boldsymbol{x}_0 = [0,0]$ and is tasked with reaching goal $B$. The predictions model an uncertain observer that holds mistaken initial beliefs $\mathcal{B}$ about the agents goals: $b_0^A=0.7$ and $b_0^B$ = 0.3. Based on these beliefs a Gaussian Mixture $p_t(\boldsymbol{x})$ is used as a prediction, with each mode assuming a Constant Velocity (CV) trajectory towards its respective goal. For timestep $t$ at each horizon step $k$: 

\begin{equation}
p_{t,k}(\boldsymbol{x}) = \sum_{g \in \mathcal{G}} b_t^g p_{t,k}(\boldsymbol{x}),
\end{equation}
where $p_{t,k}(\boldsymbol{x})=\mathcal{N}(\mu_t^g, \Sigma)$ with $\mu_{t,k}^g$ is the CV prediction for goal $g \in \mathcal{G}$ at horizon step $k$, $\Sigma$ is a fixed covariance and $\boldsymbol{x}$ is a state. We model the observer's changing beliefs $\mathcal{B}$ via Bayesian inference. With every new observation, beliefs are updated using the mode predictions $p_{t,k}(\boldsymbol{x})$ as likelihood functions:
\begin{equation}
    b_t^g = \frac{p_{t-1,0}(\boldsymbol{x}_t)b_{t-1}^g}{\sum_{g \in \mathcal{G}}p_{t-1,0}(\boldsymbol{x}_t)b_{t-1}^g }.
\end{equation}
Keeping a fixed discount $\gamma = 0.6$ in Eq. \eqref{eq:cost}, we vary the magnitude of $\lambda$ to generate results shown in Figure \ref{traj}.

\paragraph{\textbf{Results Discussion}}
We use this example to study how $\lambda$ should be tuned to control the trade-off. If predictability dominates (e.g., $\lambda = 20$ or $\lambda = 40$), this results in observations that further reinforce the observer's mistaken belief. It becomes more costly for the robot to pursue its intrinsic motivation with each time-step, thus it fails to complete the task. Conversely, if $\lambda$ is too low, the robot may still behave unpredictably\footnote{In practice, the influence seems to be very dependent on the structure of the main objective cost function, so we recommend tuning $\lambda$ empirically based on the specific planner and prediction model used.}.
For reference, the resulting behavior of an agent optimizing for \emph{legibility} as per the method of \cite{mcpherson_efficient_2021} is shown as the black line in Figures \ref{Trajs} and \ref{fig:BeliefUpdates}. From the perspective of coordination, \cite{mcpherson_efficient_2021} can be understood as an anticipatory mechanism: By conveying intention in advance, other agents anticipate better in their planning.
Our approach similarly mitigates sudden environmental changes, however instead of aiming to directly influence the other agents' beliefs, we rely on a prediction model to avoid the surprising observations throughout the interaction. While this can occasionally result in slightly more costly trajectories for the agent, we achieve similar results without requiring explicit modeling of the other agent, making it more computationally efficient and robust to situations where the agent may not be able to successfully convey its intention. As demonstrated in Figure \ref{traj}, when the observer's beliefs are misaligned, the agent adopts a pro-social behavior, gently guiding the observer toward the correct belief. 

\subsection{Robot-Robot Interactions}

\begin{figure*}[h!]
    \centering
    \subfloat[Symmetrical swapping]{\includegraphics[width=0.25\textwidth]{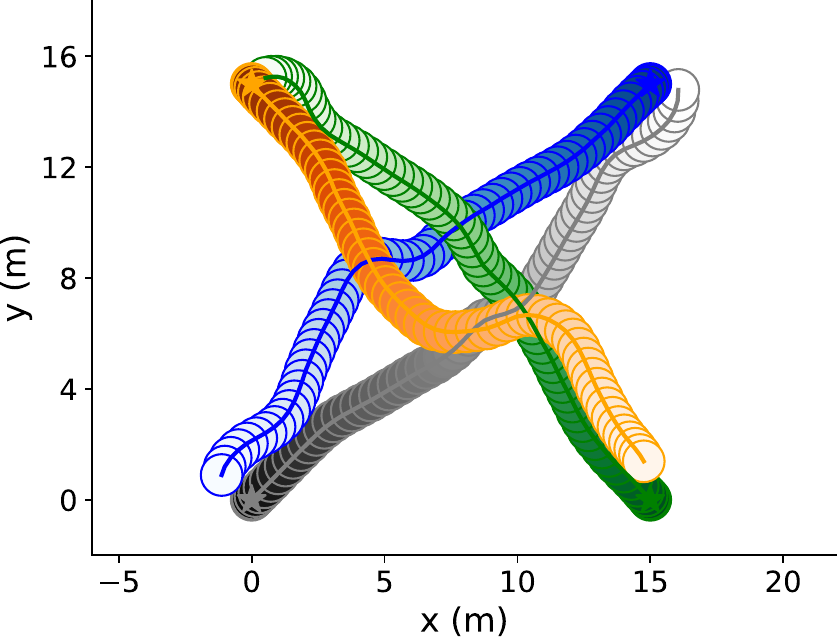}} \hspace{0.1cm}
    \subfloat[Unsymmetrical swapping]{\includegraphics[width=0.25\textwidth]{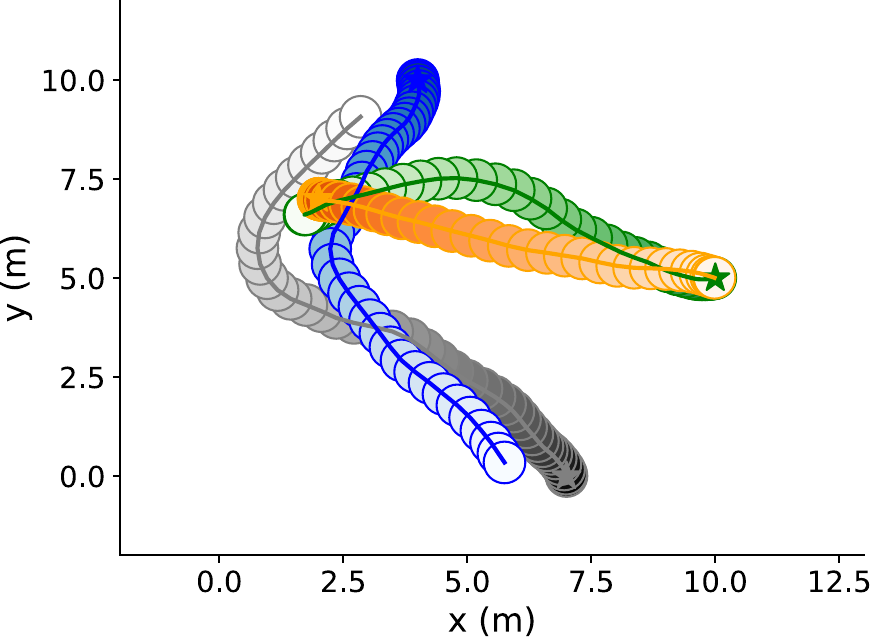}} \hspace{0.1cm}
    \subfloat[Double crossing]{\includegraphics[width=0.25\textwidth]{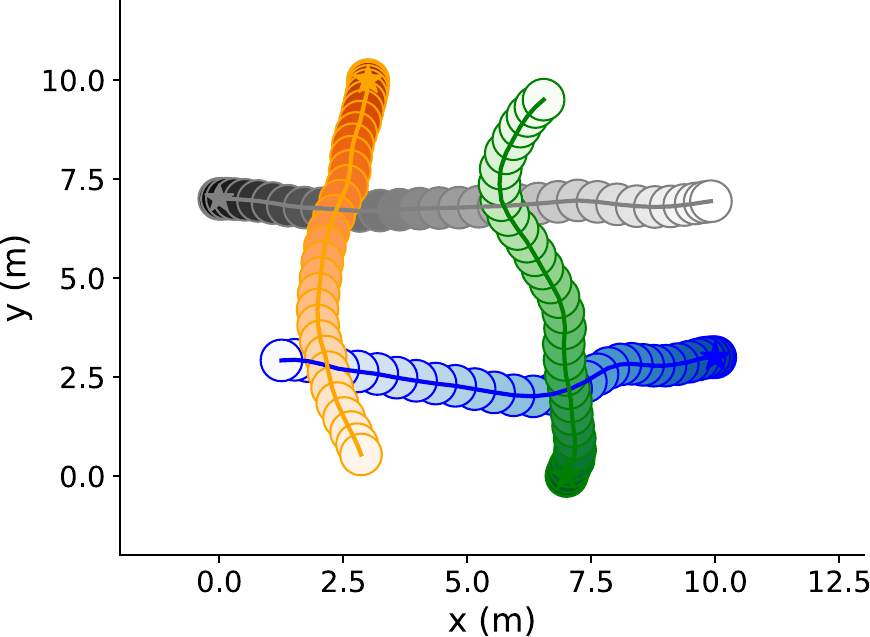}} \\
    \subfloat[Symmetrical swapping]{\includegraphics[width=0.25\textwidth]{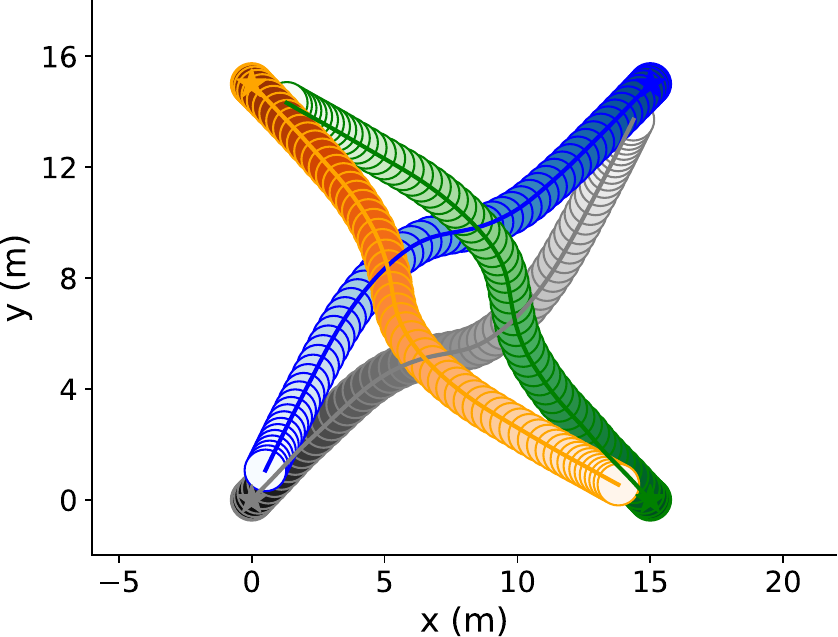}} \hspace{0.1cm}
    \subfloat[Unsymmetrical swapping]{\includegraphics[width=0.25\textwidth]{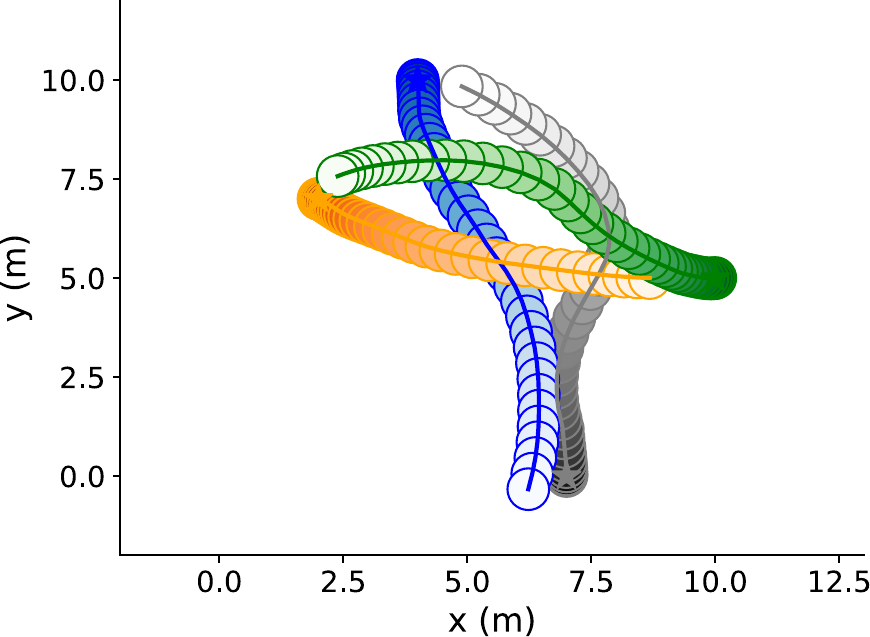}} \hspace{0.1cm}
    \subfloat[Double crossing]{\includegraphics[width=0.25\textwidth]{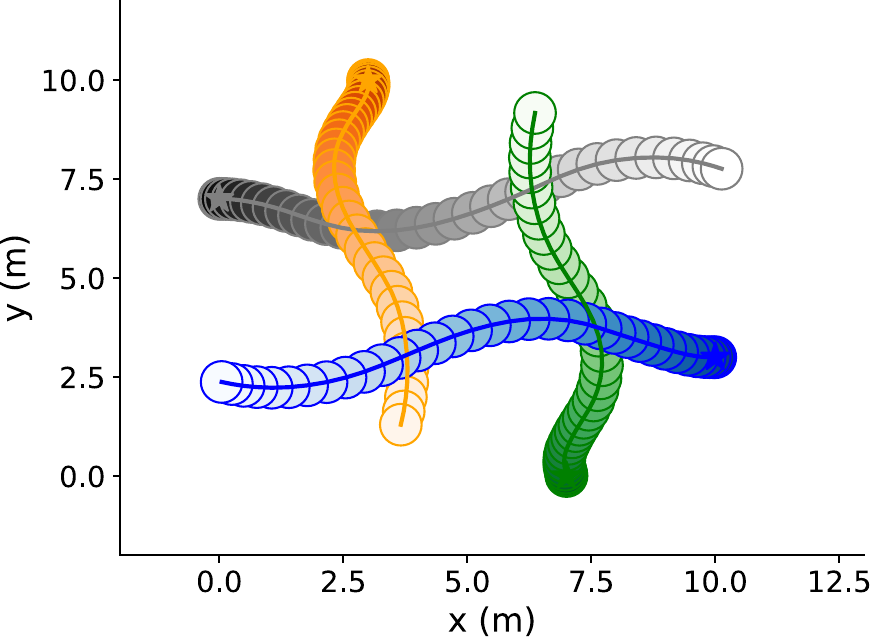}}
    \caption{The first row shows the results with $\lambda = 0$ whereas the second row shows the results for $\lambda = 5.0$. When agents account for predictability, aside from faster convergence to a coordination strategy, this also results in smoother trajectories as a consequence of better anticipation of the environment.}
    \label{Swap}
\end{figure*}

\subsubsection{Swapping Tasks}
\paragraph{\textbf{Experiment Objective}}
These experiments explore the benefits of accounting for predictability in robot-robot interactions through swapping-tasks, a common benchmark for robot coordination \cite{bhatt_efficient_2023}, \cite{zhu_learning_2021}. By performing tests in an open environment these tests avoid interference of external environmental influences.

\paragraph{\textbf{Setup}}
In the experiments, agents are initially positioned on the vertices of a square and tasked with swapping positions with the agent on the opposite vertex (Figure \ref{Swap}a). The optimal solution requires all agents to coordinate by selecting the same collision avoidance strategy, either passing left or right. Additionally, two more scenarios were tested: an asymmetrical swapping task and a double-crossing task, to explore different geometries and interactions. The experiments use a game-theoretic prediction model based on the ALGAMES framework \cite{cleach_algames_2020}, which solves constrained dynamic games to find an optimal joint strategy over a 20-step horizon. The model generates prediction distributions for each horizon step as a Gaussian with user-specified covariance $\Sigma$. By testing three values of the predictability parameter $\lambda$ \{0.0, 2.5, 5.0\}, we investigate how accounting for predictability impacts agent coordination. Each task was run 50 times, and the results for all three tasks are reported in Table \ref{CombinedTable} \footnote{For $\lambda=0$ safety constraint violations are low at 1-2 for all the tasks. For higher $\lambda$ it was 0 for all tasks. As this is not a very informative result it was not included in the tables}. An illustration comparing the trajectories for all 3 scenarios can be found in Figure \ref{Swap}.

\begin{table}[h!]
    \centering
    \caption{Table summarizing results for the 3 swapping tasks: Symmetrical, Unsymmetrical, and Double-Crossing}
    \setlength{\tabcolsep}{3pt} 
    \renewcommand{\arraystretch}{0.9} 
    \begin{tabular}{l l c c c}
    \toprule
    \textbf{Exp.} & \textbf{Metric} & \textbf{$\lambda = 0.0$} & \textbf{$\lambda = 2.5$} & \textbf{$\lambda = 5.0$} \\
    \midrule
    \multirow{3}{*}{Sym}
    & PE (m²)  & 2.116 \scriptsize{$\pm 1.000$} & 0.516 \scriptsize{$\pm 0.161$} & \textbf{0.501} \scriptsize{$\pm$ \textbf{0.145}} \\
    & Acc (m/s²) & 0.209 \scriptsize{$\pm 0.101$} & \textbf{0.038} \scriptsize{$\pm$ \textbf{0.008}} & 0.043 \scriptsize{$\pm 0.009$} \\
    & Ang (rad/s) & 0.283 \scriptsize{$\pm 0.041$} & 0.225 \scriptsize{$\pm 0.026$} & \textbf{0.219} \scriptsize{$\pm$ \textbf{0.026}} \\
    \midrule
    \multirow{3}{*}{Unsym}
    & PE (m²)  & 0.877 \scriptsize{$\pm 0.489$} & 0.291 \scriptsize{$\pm 0.178$} & \textbf{0.187} \scriptsize{$\pm$ \textbf{0.162}} \\
    & Acc (m/s²) & 0.196 \scriptsize{$\pm 0.114$} & 0.138 \scriptsize{$\pm 0.090$} & \textbf{0.112} \scriptsize{$\pm$ \textbf{0.080}} \\
    & Ang (rad/s) & 0.363 \scriptsize{$\pm 0.112$} & 0.221 \scriptsize{$\pm 0.095$} & \textbf{0.177} \scriptsize{$\pm$ \textbf{0.071}} \\
    \midrule
    \multirow{3}{*}{D-Cross}
    & PE (m²)  & 0.969 \scriptsize{$\pm 0.416$} & 0.388 \scriptsize{$\pm 0.124$} & \textbf{0.311} \scriptsize{$\pm$ \textbf{0.116}} \\
    & Acc (m/s²) & 0.249 \scriptsize{$\pm 0.124$} & \textbf{0.123} \scriptsize{$\pm$ \textbf{0.080}} & 0.125 \scriptsize{$\pm 0.070$} \\
    & Ang (rad/s) & 0.434 \scriptsize{$\pm 0.128$} & 0.283 \scriptsize{$\pm 0.096$} & \textbf{0.252} \scriptsize{$\pm$ \textbf{0.078}} \\
    \bottomrule
\end{tabular}
    \label{CombinedTable}
\end{table}

\paragraph{\textbf{Results Discussion}}
As seen in Table \ref{CombinedTable}, increasing the predictability parameter $\lambda$ consistently led to improved performance across all metrics: planning effort (PE), acceleration (Acc), and angular velocity (Ang). Notably, even selecting a small $\lambda$ causes a pronounced decrease in planning effort, with further increases in $\lambda$ yielding diminishing returns. This phenomenon can be attributed to the coordination challenge agents face in this environment, which primarily involves equilibrium selection. In situations where agents must choose between two equally viable strategies, such as passing left or passing right, our method addresses this challenge by relying on a prediction model to establish a `soft social convention'. This introduces a subtle bias towards one of the strategies, improving implicit coordination. This mechanism is particularly relevant, as prediction models often excel at capturing an agent's overarching intent and high-level strategy. However, equilibrium selection scenarios are inherently stochastic and unpredictable, making them challenging to model accurately \cite{wang_social_2022}. Thus, our method enhances robustness in such situations by guiding agents towards a coordinated strategy selected by the prediction model. In general, agents need a precise and accurate prediction model for efficient coordination. However, due to the inherent uncertainty of interactions, this is often very hard to achieve. By accounting for predictability, a group of agents is able to establish a `soft social convention' to mitigate some of this uncertainty. From the perspective of an agent, this results in more accurate predictions, allowing for smoother and more efficient coordination. This mechanism is especially effective for interactions where the main coordination challenge lies in equilibrium selection.  
\subsubsection{Robot-Robot Traffic Scenario}
\paragraph{\textbf{Experiment Objective}}
In this experiment, we focus on robot-robot coordination in driving scenarios, where the environment has a stronger influence on agent's behavior. This time, we use a data-driven prediction model to explore how predictability impacts coordination in more complex environments.

\begin{figure*}[h!]
    \centering
    \begin{subfigure}[b]{0.20\textwidth}
        \centering
        \includegraphics[width=\textwidth]{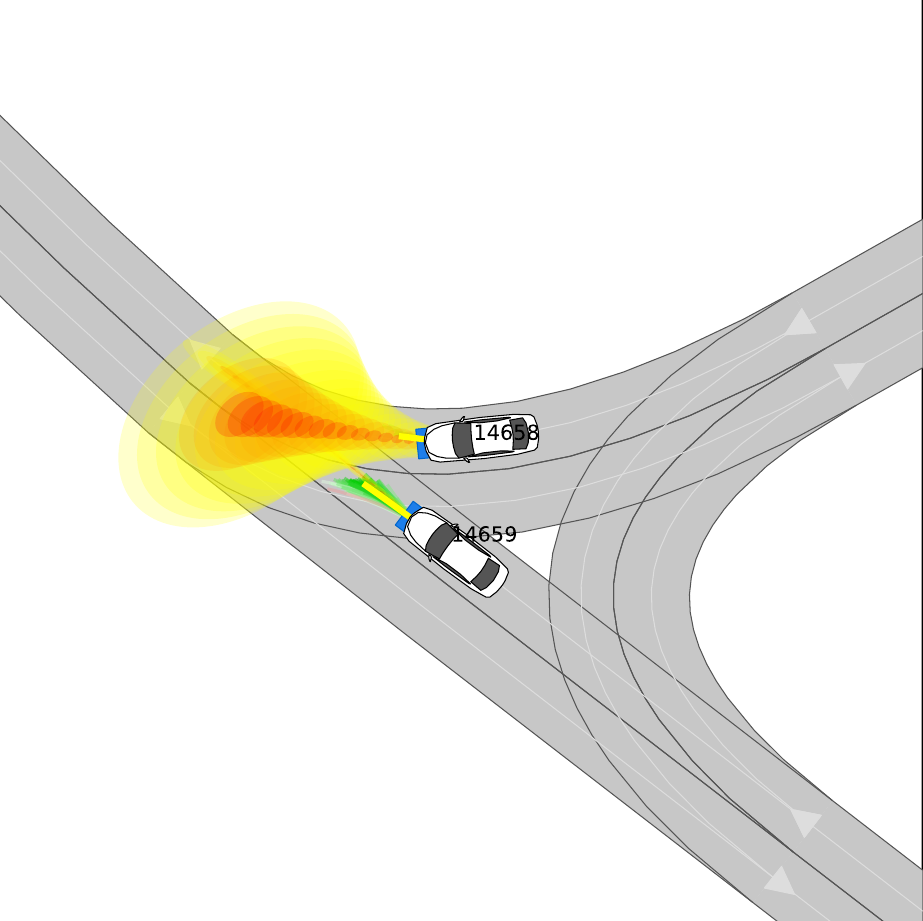}
        \caption{Lane merge $\lambda$ = 0}
        \label{fig:CRKL0}
    \end{subfigure}
    \hspace{0.01\textwidth}
    \begin{subfigure}[b]{0.20\textwidth}
        \centering
        \includegraphics[width=\textwidth]{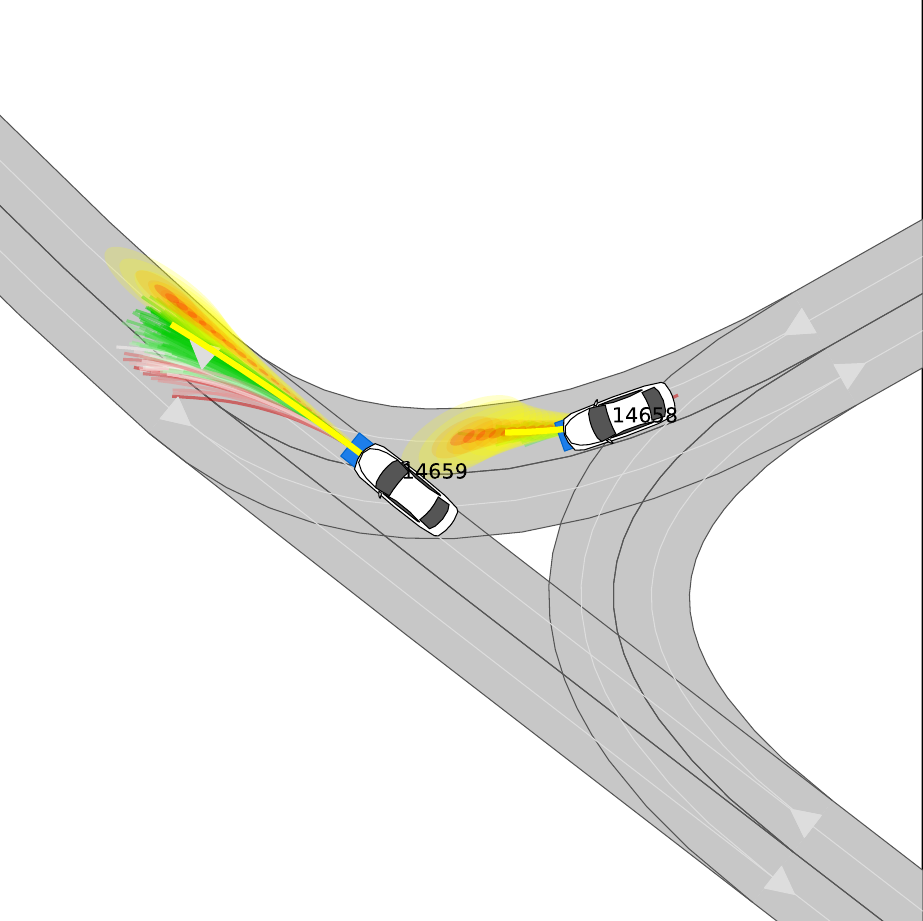}
        \caption{Lane Merge $\lambda$ = 5}
        \label{fig:CRKL5}
    \end{subfigure}
    \hspace{0.01\textwidth}
    \begin{subfigure}[b]{0.20\textwidth}
        \centering
        \includegraphics[width=\textwidth]{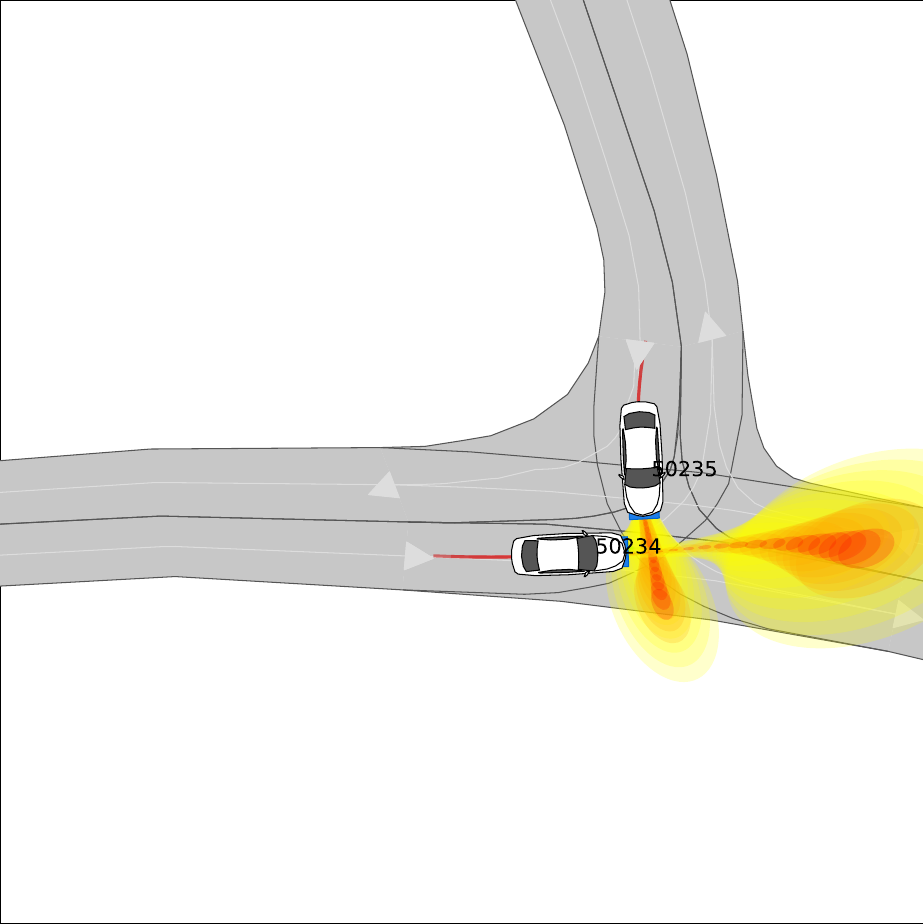}
        \caption{T-Junction $\lambda$ = 0}
        \label{fig:CrossKL0}
    \end{subfigure}
    \hspace{0.01\textwidth}
    \begin{subfigure}[b]{0.20\textwidth}
        \centering
        \includegraphics[width=\textwidth]{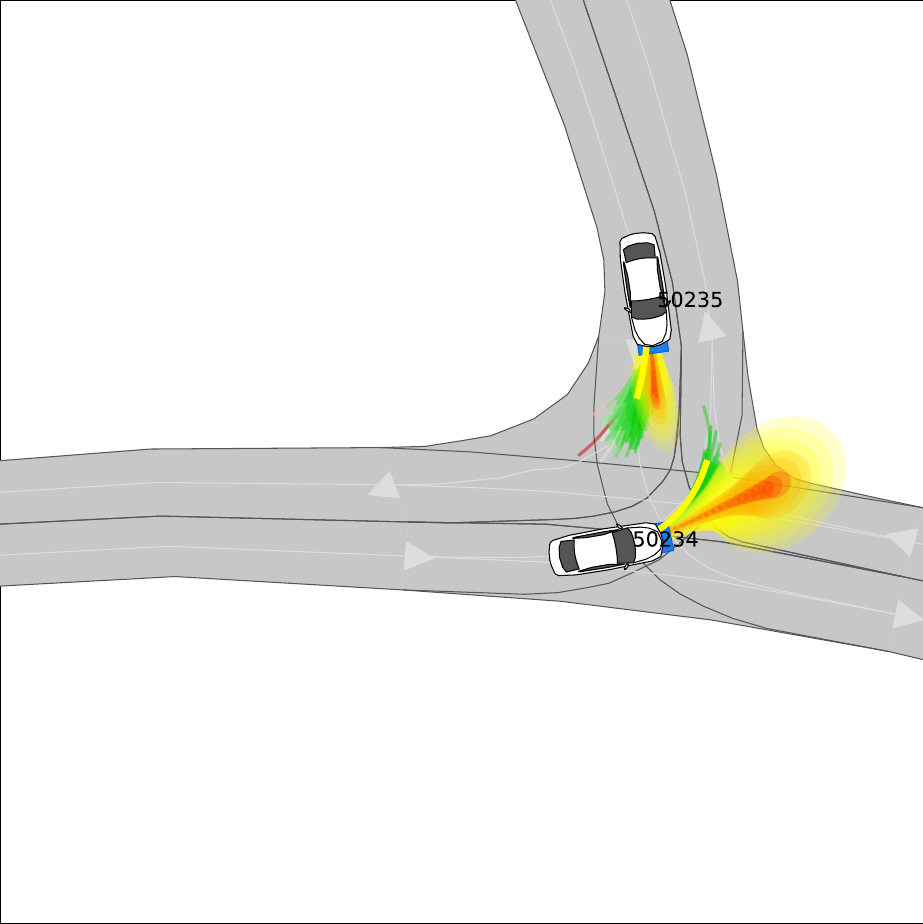}
        \caption{T-Junction $\lambda$ = 5}
        \label{fig:CrossKL5}
    \end{subfigure}
    \caption{ a) Illustration of a deadlock With $\lambda$ = 0, where a sequence of faulty predictions reinforces both agent's hesitation. b) For $\lambda$ = 5, the agents can leverage the prediction model to coordinate which agent gives way and which passes first.}
    \label{fig:CR}
\end{figure*}

\paragraph{\textbf{Setup}} We use CommonRoad \cite{althoff_commonroad_2017} as a simulator, which includes the Wale-Net \cite{geisslinger_watch-and-learn-net_2021} prediction model, a learning-based model that outputs predictions as Gaussians, accounting for uncertainty, road geometry, and the interaction with surrounding agents. Consistent with previous experiments, we employ an MPPI based planner. To account for safety in planning, we implement the constraints introduced by \cite{geisslinger_ethical_2023}, building upon and extending the code from this prior work. We perform tests in two scenarios: A T-Junction and a Lane-Merge. For both scenarios, we perform tests with $\lambda = \{0.0, 2.5, 5.0\}$ for 30 iterations applying small changes in the initial positions and velocities. An illustration of the lane merge environment is presented in Figure \ref{fig:CR}. The results for T-Junction and lane-merge are presented in Table \ref{CombinedTableScenarios}.

\begin{table}[h!]
    \centering
    \caption{Results for T-Junction and Lane Merge Scenarios (Dlk indicates Deadlocks)}
    \setlength{\tabcolsep}{3pt} 
    \renewcommand{\arraystretch}{0.9} 
    \begin{tabular}{l l c c c}
    \toprule
    \textbf{Exp} & \textbf{Metric} & \textbf{$\lambda = 0.0$} & \textbf{$\lambda = 2.5$} & \textbf{$\lambda = 5.0$} \\
    \midrule
    \multirow{5}{*}{T-J}
    & Dlk (\%) & 30.0 & \textbf{0.0} & \textbf{0.0} \\
    & Dist (m) & 30.172 & \textbf{51.248} & 47.000 \\
    & PE (m²)  & \textbf{1.366} \scriptsize{$\pm 1.126$} & 2.318 \scriptsize{$\pm 0.313$} & 2.507 \scriptsize{$\pm 0.759$} \\
    & Acc (m/s²)  & -0.142 \scriptsize{$\pm 0.238$} & \textbf{0.293} \scriptsize{$\pm 0.045$} & 0.287 \scriptsize{$\pm 0.233$} \\
    & Ang (rad/s)  & 0.0037 \scriptsize{$\pm 0.0028$} & \textbf{0.0005} \scriptsize{$\pm 0.0002$} & 0.0028 \scriptsize{$\pm 0.0026$} \\
    \midrule
    \multirow{5}{*}{LM}
    & Dlk (\%) & 73.3 & \textbf{0.0} & \textbf{0.0} \\
    & Dist (m) & 46.878 & \textbf{75.800} & 69.909 \\
    & PE (m²)   & \textbf{2.079} \scriptsize{$\pm 0.785$} & 3.513 \scriptsize{$\pm 0.564$} & 3.315 \scriptsize{$\pm 0.760$} \\
    & Acc (m/s²)  & \textbf{0.111} \scriptsize{$\pm 0.092$} & 0.337 \scriptsize{$\pm 0.054$} & 0.317 \scriptsize{$\pm 0.087$} \\
    & Ang (rad/s)  & 0.0032 \scriptsize{$\pm 0.0032$} & 0.0009 \scriptsize{$\pm 0.0006$} & \textbf{0.0001} \scriptsize{$\pm 0.0001$} \\
    \bottomrule
\end{tabular}
    \label{CombinedTableScenarios}
\end{table}
\paragraph{\textbf{Results Discussion}}

When agents fail to coordinate in road scenarios, they often experience deadlocks or, in the worst case, collisions. In Figure \ref{fig:CRKL0}, an example of a deadlock is illustrated. Deadlocks are common in limited space environments such as intersections or narrow passages. Initially, the model may predict one agent will yield while the other advances. However, as  deviations occur and both agents hesitate, their predictions begin to reinforce each other’s hesitation, creating the deadlock. The model may then be unable to introduce asymmetry to prioritize one of the agents in ambiguous situations, preventing the agents from breaking away from the deadlock. Results show that agents incorporating predictability into their models achieve better coordination, as indicated by less pronounced slowdowns resulting in higher traveled distance and the \emph{disappearance of deadlocks} as seen in Table \ref{CombinedTableScenarios}. When examining other metrics, the benefits of incorporating predictability are not as pronounced, especially for higher $\lambda$. This occurs because the prediction model is not explicitly conditioned to align with the road geometry (Figures \ref{fig:CrossKL0},\ref{fig:CrossKL5}). Since the planner is required to track a reference path, deviations between the predictions and the reference path can push the agent to deviate from the path, requiring small adjustments more frequently for higher higher $\lambda$. Although this problem has marginal impact on the overall performance of the agent, the reduction in planning effort may be mitigated. 

Similar to the Swapping Task tests, we see that agents are able to use the prediction model to coordinate by reducing uncertainty on equilibrium selection, namely, which agent gives way. However, a noteworthy observation is that, beyond reducing uncertainty, agents enhance their performance by adopting pro-social behaviors embedded in the model's latent space. These behaviors include adherence to social norms and subtle cues learned from training data, mirroring the behavior of experts used to train the model. This behavior resembles imitation learning, where agents learn cooperative strategies directly from expert demonstrations embedded in the prediction model. As seen in Figure \ref{fig:CRKL5}, although both outcomes are equally plausible from the raw planning problem, agents consistently converge on the solution where the merging agent yields, which aligns with typical human driving patterns.

\subsection{Experiments with human-driver data}
\begin{figure*}[h!]
    \centering
    \begin{minipage}{0.3\textwidth}
        \centering
        \includegraphics[width=0.7\textwidth]{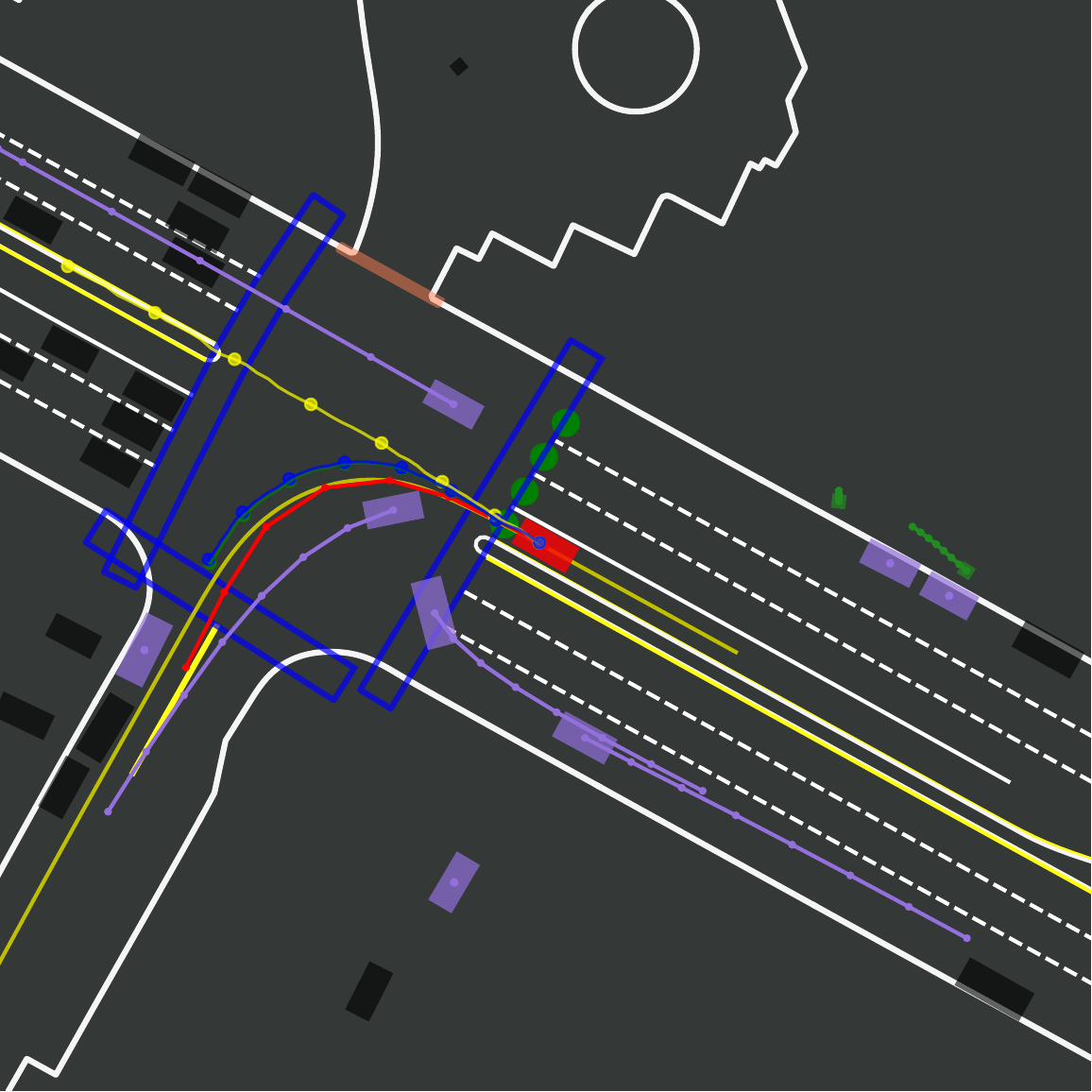}
        \caption{Illustration of the navigation problem in Crossing1. The reference global path is rendered as a smooth yellow line. The AV's plan is rendered in red. Predictions for other agents are rendered in purple, showing only the most likely mode for clarity. The ego-prediction is multi-modal with 3 modes represented by the yellow, green, and blue trajectories.}
        \label{fig:SceneDIPP}
    \end{minipage}%
    \hfill
    \begin{minipage}{0.65\textwidth}
        \centering
        \captionof{table}{Results comparing the performance of an MPPI-based planner on Waymo Open Motion Dataset scenarios for different $\lambda$ values. For 30 iterations we present the number of collisions and the mean value of other performance metrics}
        \setlength{\tabcolsep}{3pt} 
    \renewcommand{\arraystretch}{0.9} 
\begin{tabular}{l l r r r r r}
    \toprule
    \textbf{Scenario} & \textbf{$\lambda$}  & \textbf{Col (\%)} & \textbf{Dist (m)} & \textbf{Acc (m/s²)} & \textbf{Lat\_Acc (m/s²)} & \textbf{L2 (m)} \\ 
    \midrule
    \multirow{3}{*}{Crossing1} 
    & $ 0$   & 43.3 & 74.540 \scriptsize{$\pm 1.928$} & 1.085 \scriptsize{$\pm 0.121$} & 1.615 \scriptsize{$\pm 0.578$} & 4.101 \scriptsize{$\pm 0.532$} \\
    & $ 75$  & 0 & 68.353 \scriptsize{$\pm 0.982$} & 0.681 \scriptsize{$\pm 0.025$} & 0.412 \scriptsize{$\pm 0.044$} & 3.358 \scriptsize{$\pm 0.221$} \\
    & $ 120$ & 0 & 55.796 \scriptsize{$\pm 1.321$} & 0.472 \scriptsize{$\pm 0.035$} & 0.149 \scriptsize{$\pm 0.025$} & 2.554 \scriptsize{$\pm 0.053$} \\
    \midrule
    \multirow{3}{*}{Crossing2} 
    & $ 0$   & 86.6 & 75.843 \scriptsize{$\pm 4.803$} & 1.418 \scriptsize{$\pm 0.099$} & 2.121 \scriptsize{$\pm 0.295$} & 12.707 \scriptsize{$\pm 1.112$} \\
    & $ 75$  & 0 & 55.353 \scriptsize{$\pm 1.403$} & 0.957 \scriptsize{$\pm 0.091$} & 0.314 \scriptsize{$\pm 0.036$} & 4.603 \scriptsize{$\pm 1.246$} \\
    & $ 120$ & 23.3 & 37.518 \scriptsize{$\pm 12.954$} & 1.534 \scriptsize{$\pm 0.736$} & 0.227 \scriptsize{$\pm 0.084$} & 2.947 \scriptsize{$\pm 3.206$} \\
    \midrule
    \multirow{3}{*}{Intersection} 
    & $ 0$   & 26.6 & 69.747 \scriptsize{$\pm 4.794$} & 1.450 \scriptsize{$\pm 0.153$} & 1.817 \scriptsize{$\pm 0.782$} & 24.808 \scriptsize{$\pm 3.632$} \\
    & $ 75$  & 0 & 72.700 \scriptsize{$\pm 0.115$} & 0.709 \scriptsize{$\pm 0.029$} & 0.493 \scriptsize{$\pm 0.075$} & 24.618 \scriptsize{$\pm 1.036$} \\
    & $ 120$ & 0 & 71.885 \scriptsize{$\pm 0.227$} & 0.613 \scriptsize{$\pm 0.043$} & 0.304 \scriptsize{$\pm 0.026$} & 19.900 \scriptsize{$\pm 0.733$} \\
    \midrule
    \multirow{3}{*}{Emergency} 
    & $ 0$   & 53.3 & 61.377 \scriptsize{$\pm 20.244$} & 1.258 \scriptsize{$\pm 0.175$} & 0.882 \scriptsize{$\pm 0.141$} & 8.631 \scriptsize{$\pm 5.877$} \\
    & $75$  & 0 & 68.883 \scriptsize{$\pm 0.525$} & 0.763 \scriptsize{$\pm 0.010$} & 0.365 \scriptsize{$\pm 0.031$} & 1.961 \scriptsize{$\pm 0.069$} \\
    & $120$ & 0 & 60.058 \scriptsize{$\pm 0.797$} & 0.581 \scriptsize{$\pm 0.019$} & 0.184 \scriptsize{$\pm 0.016$} & 1.515 \scriptsize{$\pm 0.091$} \\
    \bottomrule
\end{tabular}

        \label{DIPPTable}
    \end{minipage}
\end{figure*}

\paragraph{\textbf{Experiment Objective}}
The goal of this experiment is to evaluate whether predictability can bridge the gap between algorithmic planning and the natural driving patterns observed in humans, facilitating smoother and more adaptive interactions in complex driving environments. We test this by incorporating predictability with simple MPPI-based reference-tracking planner using a SOTA data-driven prediction model.

\paragraph{\textbf{Setup}} We utilize a state-of-the-art (SOTA) prediction model introduced by \cite{huang_differentiable_2023}, a multi-modal, transformer-based architecture trained on the Waymo Open Motion Dataset. The model generates scene-centric predictions with three modes, representing the most likely joint trajectories of up to 11 agents, including the ego agent. An MPPI planner is used for reference tracking, incorporating collision avoidance as outlined in \cite{huang_differentiable_2023}. For this experiment, we replay recorded scenes from the Waymo dataset's test set, meaning agents in the environment follow pre-recorded, non-interactive trajectories. The goal is for the ego agent to replicate expert behavior observed during training. We perform tests for $\lambda = {0, 75, 120}$\footnote{The large $\lambda$ values here respond to the particular magnitude of the planning cost function and the prediction model used. We found that values of a higher order of magnitude were needed to obtain predictable behavior shifts.}, over 30 iterations in selected scenarios that require human-like interactions, similar to the approach of \cite{huang_differentiable_2023}. A screenshot of the crossing scenario is shown in Figure \ref{fig:SceneDIPP}, and results are reported in Table \ref{DIPPTable}.


\paragraph{\textbf{Results Discussion}} From Table \ref{DIPPTable}, it is evident that increasing the weight of the predictability objective results in fewer collisions and smoother control inputs. Interestingly, however, this does not necessarily correlate with improved progress along the reference path. This can be attributed to the planner inducing less distributional shift in the prediction model. The model, trained on scenes where all agents exhibit expert behaviors, struggles when the planner deviates significantly from these patterns, as it encounters situations outside its training distribution. In such cases, the model attempts to extrapolate and produces sub-optimal predictions, such as incorrectly anticipating that an agent may yield or maneuver differently than it actually does based on the recorded data. This misalignment leads to overconfident behavior in some instances, which, while promoting progress along the reference path, increases the risk of collisions. Evidence supporting this hypothesis is found in the Human L2 loss metric, which measures the L2 loss between the agent's trajectory and the corresponding human trajectory that the planner aims to replicate. For $\lambda=0$, the higher L2 loss indicates significant deviation from human behavior, suggesting that the agent diverges more from the expert’s trajectory. In contrast, when predictability is considered, the L2 loss decreases, indicating that the agent’s behavior aligns more closely with the human data. This results in reduced distributional shift and, consequently, more accurate predictions and smoother trajectories.


\section{Discussion}

\paragraph{\textbf{Discussion}} The method assumes that agents can approximate each other's expectations, often implying a shared prediction model. Although this might seem impractical, certain decentralized settings could accommodate shared models. For instance, in a warehouse environment where multiple Autonomous Ground Vehicles (AGVs) transport valuable goods, a shared prediction model could be feasibly developed and implemented \cite{zhu_learning_2021}. When integrated with our methodology, such a model could establish 'operational norms', enabling agents to coordinate efficiently and robustly without the need for centralized control, thus reducing computational and infrastructure demands. A comparable scenario is anticipated in future markets where autonomous vehicles (AVs) from different manufacturers must interact. Recent studies pointed at the importance of establishing a unified driving convention \cite{yu_will_2022}, as the absence of such a standard could lead to exploitative strategies from different AV companies pursuing competitive advantage, and thereby compromise safety.  Different companies can cooperate to develop a shared prediction model to serve as an industry standard. Given a model all AVs in traffic share, our method would enable AVs to anticipate each other's actions and more effectively settle on coordination, thereby providing enhanced road safety without explicit coordination or reliance on infrastructure for centralized coordination.

\paragraph{\textbf{Future Work}} Balancing predictability and performance cost, determined by $\lambda$, is complex and context-dependent. Dynamically adjusting $\lambda$ as agents interact could improve performance, increasingly prioritizing predictability in safety-critical moments. Developing adaptive heuristics for this adjustment, as suggested by previous work \cite{bastarache_legible_2023,dragan_integrating_2014}, would be a valuable research direction. Alternatively, using lexicographic optimization could enhance generalizability by providing a structured trade-off that eliminates the need for tuning a magnitude dependent weighting parameter. However, This requires adpatation of the cost function computation to account for the lexicographic priorities, where predictability is prioritized subject to a performance constraint.



%

%

\paragraph{\textbf{Conclusion}} We present a novel approach to enhance multi-agent interaction capabilities for sequential predict-and-plan frameworks by introducing predictability as a key optimization objective. Accounting for predictability in this manner can be understood as an implicit cooperation mechanism whereby agents use a prediction model to actively reduce uncertainty about the environment for other agents. This not only improves the robustness of coordination strategies but also reduces planning effort without requiring explicit communication or high-level control, and does so independently of the number of interacting agents. Through experiments, including robot-robot interactions and human-interaction scenarios, our method improved agent coordination, reduced collisions, and led to smoother, more efficient trajectories (particularly in complex coordination environments). We also demonstrated that the benefits extend to interactions with human drivers by allowing the agent to more reliably use its prediction model.

\label{Conclusion}




\bibliographystyle{ACM-Reference-Format} 
\bibliography{paper}


\end{document}